\documentclass[10pt,twocolumn,letterpaper]{article}

\usepackage[pagenumbers]{iccv} 
\usepackage{amsfonts}  
\usepackage{graphicx}
\usepackage{colortbl}
\usepackage{makecell}
\usepackage{amssymb}
\usepackage{pifont}
\usepackage{array} 
\usepackage{float}
\usepackage{multirow}
\usepackage{url}
\definecolor{iccvblue}{rgb}{0.21,0.49,0.74}
\usepackage[pagebackref,breaklinks,colorlinks,allcolors=iccvblue]{hyperref}

\def\paperID{8001} 
\def\confName{ArXiv}
\def\confYear{2025}

\title{SS-DC: Spatial-Spectral Decoupling and Coupling Across Visible-Infrared Gap for Domain Adaptive Object Detection}

\author{
Xiwei Zhang\textsuperscript{1}, Chunjin Yang\textsuperscript{1}, Yiming Xiao\textsuperscript{1}, Runtong Zhang\textsuperscript{1}, Fanman Meng\textsuperscript{1}\textsuperscript{,*} \\ 
\textsuperscript{1}University of Electronic Science and Technology of China,\\ Chengdu, Sichuan 611731, China \\
{\tt\small \{202322011832, 202322011830, ymxiao, 202211012322\}@std.uestc.edu.cn}
{\tt\small ,\textsuperscript{*}fmmeng@uestc.edu.cn}              
}

\begin{document}
\maketitle
\begin{abstract}

    Unsupervised domain adaptive object detection (UDAOD) from the visible domain to the infrared (RGB-IR) domain is challenging. Existing methods regard the RGB domain as a unified domain and neglect the multiple subdomains within it, such as daytime, nighttime, and foggy scenes. We argue that decoupling the domain-invariant (DI) and domain-specific (DS) features across these multiple subdomains is beneficial for RGB-IR domain adaptation. To this end, this paper proposes a new SS-DC framework based on a decoupling-coupling strategy. In terms of decoupling, we design a Spectral Adaptive Idempotent Decoupling (SAID) module in the aspect of spectral decomposition. Due to the style and content information being highly embedded in different frequency bands, this module can decouple DI and DS components more accurately and interpretably. A novel filter bank-based spectral processing paradigm and a self-distillation-driven decoupling loss are proposed to improve the spectral domain decoupling. In terms of coupling, a new spatial-spectral coupling method is proposed, which realizes joint coupling through spatial and spectral DI feature pyramids. Meanwhile, this paper introduces DS from decoupling to reduce the domain bias. Extensive experiments demonstrate that our method can significantly improve the baseline performance and outperform existing UDAOD methods on multiple RGB-IR datasets, including a new experimental protocol proposed in this paper based on the FLIR-ADAS dataset.
    
\end{abstract}    
\section{Introduction}
\label{sec:intro}


\begin{figure}[t]
  \centering
   \includegraphics[width=1.0\linewidth]{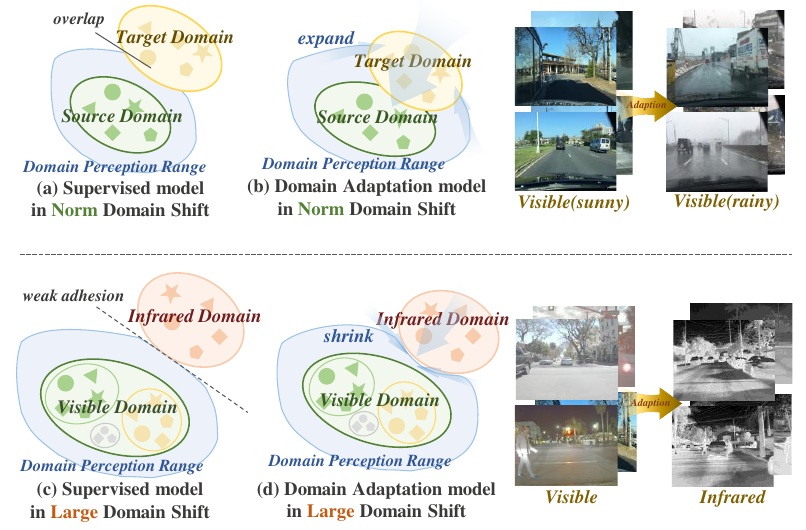}

   \caption{(a) and (b) show the impact of UDAOD methods for weather change within the RGB domain on the data distribution. (c) and (d) show the challenges of conventional UDAOD methods in RGB-IR domain adaptation which lead the model to deviate from the target domain during training process due to misalignment or unstable/incorrect pseudo-labels.}
   \label{fig:intro}
\end{figure}


In practical applications, labeled infrared data is extremely scarce, which severely limits the performance of object detection in infrared domain \cite{akshatha2022human,lu2021bridging,ye2023channel,chen2021neural,gundougan2023ir}. At the same time, there are a sufficient number of RGB images with annotations in reality \cite{lin2014microsoft,everingham2015pascal,krasin2017openimages,shao2019objects365}. These RGB datasets can be used to train object detection models and, through domain adaptation, to address the object detection problem in infrared domain \cite{Hao2024SimplifyingSD,khanh2024dynamic,liu2023periodically,Danish_2024_CVPR}.


Unsupervised Domain Adaptive Object Detection (UDAOD) methods \cite{he2024diffusion,yang2024rethinking,cao2023contrastive,chen2025datr,deng2023harmonious,feng2023dsd,li2022cross,zhao2023masked} have made significant progress in adapting from labeled source domains to unlabeled target domains within the RGB domain, typically using mean-teacher frameworks or cross-domain feature alignment. However, these methods face critical challenges as illustrated in Figure \ref{fig:intro}. Specifically, challenges include: a) \textit{Pseudo-label reliability degradation}: Teacher-student paradigms often generate invalid pseudo-labels due to inappropriate knowledge, which misguide the student model's training and trigger catastrophic performance degradation; b) \textit{Feature alignment ineffectiveness}: The significant disparity between RGB and IR domains hinders feature alignment strategies from achieving alignment without compromising the convergence stability of detection loss. 

A small yet innovative subset of recent studies \cite{do2024d3t} have explored specialized adaptation designs for RGB-IR domain shifts, achieving innovative and promising results. However, these approaches mostly neglect to account for the muti-subdomain distribution of RGB domain. The DI and DS information among subdomains in the RGB domain could provide knowledge for the UDAOD learning process from RGB to IR domains. Hence, we naturally raise an open research question: \textit{how to effectively leverage the complex data distribution within the RGB domain to enhance the adaption to the IR domain?} Spectral decomposition has been long acknowledged effective to handle the style and content information separately in the context of domain adaption \cite{Bo2024FAMNetFM, yi2024learning, NEURIPS2024_d3b8ce5e, DBLP:journals/corr/abs-2406-02833}, where the style and content information are highly embedded in the high and low frequency components. That leads our further question: \textit{how to construct a spectrum-based method to decouple DI and DS information from RGB sub-domains and apply it to the IR domain?}

In this paper, we propose a Spatial-Spectral Decoupling and Coupling (SS-DC) framework based on the Mean-Teacher \cite{liu2021unbiased, tarvainen2017mean,dobler2023robust} and Detection Transformer (DETR) \cite{carion2020end,zhang2022dino,li2022dn}. In the SS-DC framework, we first construct the Spectral Adaptive Idempotent Decoupling (SAID) module in the frequency domain branch. SAID includes an adaptive Filter Bank-based spectral processing paradigm and a self-distillation-driven decoupling loss design. In the Burn-in stage of the Mean Teacher, SAID leverage the multi-subdomain distribution of RGB to enables the extraction of DI information and DS interference from spectral features. Moreover, this strong decoupling capability can make the initialized model less sensitive to domain bias when generating pseudo-labels in next stage and be further enhanced in the Teacher-Student mutual learning stage. This decoupling expression does not force the alignment of the original features of the source and target domains, thereby independently enhancing the adaptability of UDAOD from the RGB domain to the IR domain.

On this basis, we propose the Spatial-Spectral Coupling (SS-Coupling) method to enhance DI information by realizing joint coupling through spatial and spectral feature pyramid in different manners tailored to different semantic levels. The SS-Coupling method also embeds DS interference as a set of tokens and feeds them into the Transformer's Encoder \cite{vaswani2017attention} to guide the model in adaptively reducing the DS interference between RGB and IR domains.

Underpinned by these designs, the detection performance in the UDAOD task with a large domain gap is markedly improved. In summary, our contributions are as follows:

\begin{itemize}
  \item We propose the SS-DC framework that leverages both the spatial and spectral domains, and uses the dual-branch network to enhance the UDAOD process.
  \item We utilize the SAID module to perform self-distillation-driven decoupling of spectral information and explore a novel paradigm for frequency-domain processing.
  \item We use the SS-Coupling method to interact the decoupled spectral features with spatial features in a multi-level, multi-structured way; this effectively bridges the domain gap between RGB and IR domains.
  \item We make a new experimental protocol based on the FLIR-ADAS \cite{farooq2021object} dataset to more realistically verify the adaptation effect from the RGB source domain to the IR target domain, and demonstrate the effectiveness of our method.
\end{itemize}

\section{Related Work}
\subsection{Self-training framework}
Self-training frameworks have been extensively adopted in cross-domain adaptation tasks. These frameworks \cite{liu2021unbiased,xu2021end,li2022cross,zhang2022dino} utilize unlabeled data to generate pseudo-labels for semi-supervised training, significantly minimizing dependence on costly manual annotations and enhancing the adaptation. LabelMatch \cite{chen2022label} introduced dynamic filtering thresholds to refine pseudo-labels. Semi-DETR \cite{zhang2023semi} adapted Mean Teacher \cite{tarvainen2017mean} to DINO-DETR \cite{zhang2022dino}, enhancing stability via query exchange. \cite{cai2019exploring,deng2021unbiased,li2022cross,deng2021unbiased,chen2024cross} have demonstrated the efficacy of the Mean Teacher framework in cross-domain scenarios, where its integration with domain alignment learning effectively mitigates domain shifts.

\subsection{UDAOD against RGB-subdomains}
Current UDAOD methods mainly fall into two categories: self-learning and alignment-based approaches, which often work together to accomplish UDAOD tasks. Self-learning methods adapt different domains by iteratively optimizing target-domain pseudo-labels. CMT \cite{cao2023contrastive} integrates contrastive learning into self-training without target annotations; and HT \cite{deng2023harmonious} improves consistency through sample reweighting. Alignment-based methods reduce domain gaps by aligning feature distributions across domains at four levels: Image-level alignment \cite{he2024diffusion} adjusts global style distributions in input or shallow features. Feature-level alignment \cite{feng2023dsd,han2024remote} uses adversarial training to extract domain-invariant features. Instance-level alignment \cite{xu2022h2fa,he2023bidirectional} matches instance features from RPN or detection heads. Category-level alignment \cite{li2022sigma,chen2025datr} optimizes semantic coherence via meta-learning or knowledge bases. However, these methods face challenges in RGB-IR adaptation. Self-learning relies on pseudo-label quality, which can degrade when RGB and IR data have significant discrepancies, leading to misguidance in the learning process. Alignment-based methods struggle to achieve effective alignment when the internal distribution of the RGB domain is complex and diverges significantly from the IR domain.

\subsection{UDAOD for RGB-IR}
Due to the significant domain gaps between RGB and IR domains, traditional UDAOD methods struggle to achieve satisfactory results. Despite its practical importance, research in this field remains limited. Meta-UDA \cite{vs2022meta} optimizes RGB-IR adaptation via an algorithm-agnostic meta-learning framework, marking a breakthrough. CutMix-based \cite{nakamura2022few} data fusion strategy, integrates RGB image patches into IR domains and boosting detection performance with adversarial learning. Dinh et al. D3T \cite{do2024d3t} framework proposes a novel Zigzag learning mechanism tailored for RGB-IR adaptation, enhancing adaptation capabilities through dual-teacher guidance. Current challenges include extracting shared information under large RGB-IR domain gaps and leveraging RGB domain internal variability. Existing methods lack explicit analysis of intra-domain differences in RGB data. To address this, we propose the SS-DC framework, which explicitly decouples domain-invariant information to significantly improve RGB-to-IR adaptation.

\section{Methodology}

\subsection{Problem Formulation}

Let $\mathcal{D}_s = \{ \mathcal{D}_{s_1}, \mathcal{D}_{s_2}, ..., \mathcal{D}_{s_K} \}$ denote the RGB source domain, where each sub-domain $\mathcal{D}_{S_k} = \{ (x^i_{s_k} , y^i_{s_k} ) \}_{i=1}^{N_k}$ includes $N_k$ images $\{x^i_{s_k}\}_{i=1}^{N_k}$ and the corresponding label $\{y^i_{s_k}\}_{i=1}^{N_k}$ in \textit{k-th} sub-domain. $D_t=\{{x^j_t}\}_{j=1}^{N_t}$ denotes unlabeled IR target domain, which consists of $N_t$ images. Specifically, ${D}_{S_k}$ is following the distribution $P_{S_k}(x,y)$ , containing bounding box coordinates $b_i^{S_k} 
 \in \mathbb{R}^4$  and instance category labels $c_i^{S_k} \in \mathcal{O}$ (where $\mathcal{O} ={o_1,...,o_M}$ is the instance-level semantic space). $D_t$ is following the distribution $P_t(x)$, whose instance semantic space is consistent with the source domain. The primary objective in RGB$\rightarrow$IR is to get a decoupling function $\phi:\mathcal{X} \rightarrow \mathcal{H}, \mathcal{H} = \{\mathcal{H}_{inv},\mathcal{H}_{spe}\} $ , which projects the input data $\mathcal{X}$ into a decoupling space $\mathcal{H}$, and independently utilizing $\mathcal{H}$ to enhance the performance of domain-adaptation from RGB to IR domains without forced alignment.


\subsection{Framework Overview}

\begin{figure*}[h]
  \centering
   \includegraphics[width=1\linewidth]{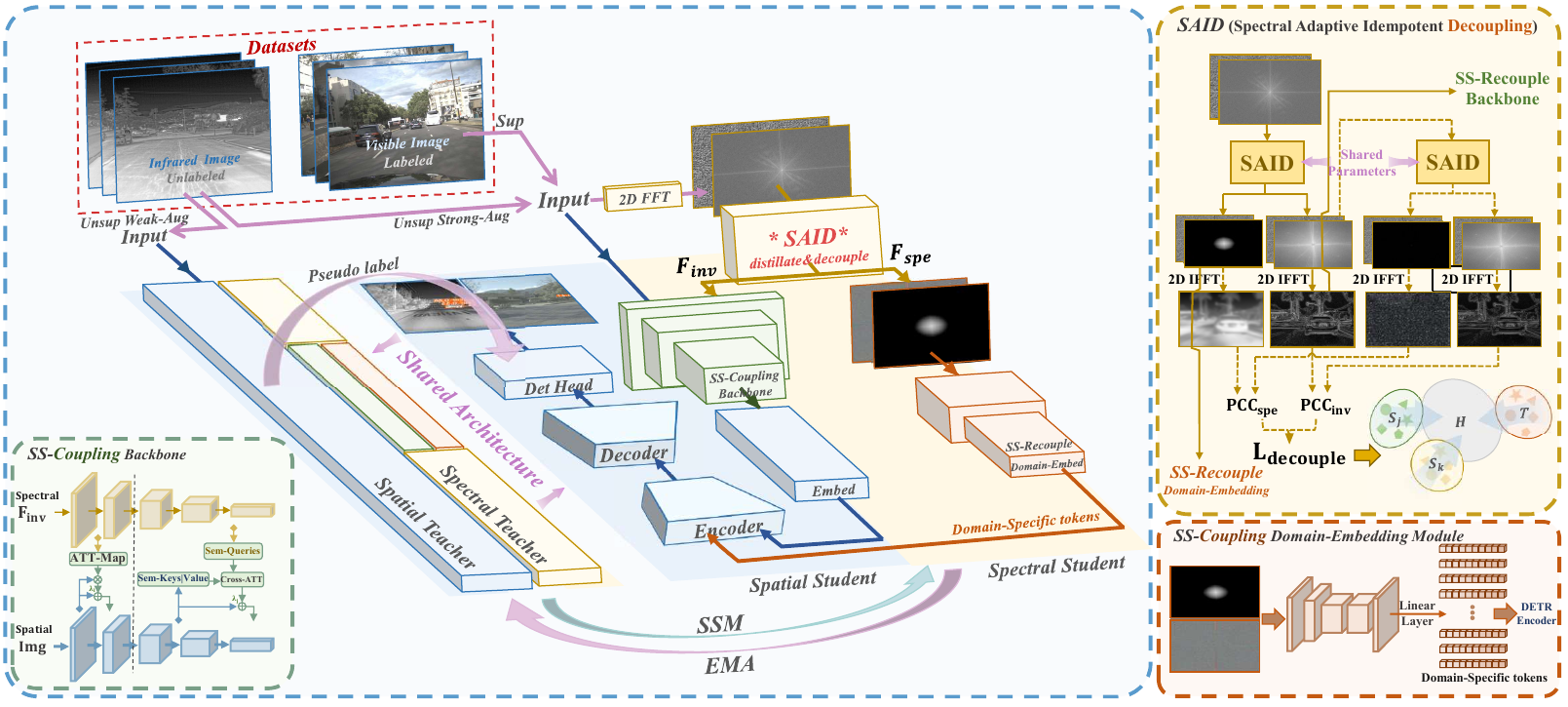}

   \caption{The SS-DC framework is based on the Mean Teacher framework. The Spectral Student uses the SAID module for self-distillation and frequency-domain decoupling, separating spectral information into DI and DS components. Both students are coupled through various strategies to enhance domain adaptation. Training is divided into two stages: \textbf{Burn-in stage}, where the model learns basic domain adaptation from RGB domain variation. \textbf{Teacher-Student mutual learning stage}, where pseudo-labels generated by the teacher model update the student model using unlabeled IR data. The teacher model is updated via EMA from the student model, and we further stabilize the adaptation process with the SSM method.}
   \label{fig:framework}
\end{figure*}

As shown in Figure \ref{fig:framework}, the SS-DC framework is trained under the Mean Teacher semi-supervised paradigm based on DETR framework. The teacher model and the student model share the same network architecture. The teacher model is updated via Exponential Moving Average (EMA) \cite{hunter1986exponentially} (Sec. \ref{Sec.MT}). Input images are decomposed into amplitude and phase spectra via 2D Fast Fourier Transform \cite{nussbaumer1982fast}. The spectral information undergoes decoupling through the idempotent filtering layer, separated into DI information and DS interference. A self-distillation-driven decoupling constraint is used to ensure decoupling stability (Sec. \ref{sec.SAID}). The decoupled spectral features is deeply coupled with the spatial features in a multi-level, multi-structured way(Sec. \ref{sec.Couple}).
This allows interaction of features between RGB domain and IR domain, thereby bridging the significant domain gap between the RGB domain and the IR domain.

\subsection{Mean Teacher Training on DETR Detector}
\label{Sec.MT}

The Mean Teacher framework exhibits strong robustness to noise in domain adaptation tasks through dynamic parameter aggregation and consistency regularization. DETR\cite{carion2020end}, with its end-to-end detection architecture and global attention mechanism, effectively captures the contextual dependencies of cross-domain targets. On one hand, DETR's set prediction characteristics mitigate the redundant matching problem in cross-domain pseudo-labels; on the other hand, the EMA mechanism in Mean Teacher provides a stable optimization path for RGB-IR cross-domain knowledge adaptation. The training process consists of two stages:

\textbf{Source Domain Supervised Burn-in stage}: The model undergoes fully supervised training based on labeled RGB data $\mathcal{D}_s$ from the source domain with $\mathcal{L}_{sup}$. In addition, we utilize a decoupling loss $\mathcal{L}_{\mathrm{dcp}}$ (Sec. \ref{sec.SAID}) to constrain the model, enforcing a strong decoupling of the frequency-domain input into DI and DS components. 
\begin{equation}
\begin{aligned}
 \mathcal{L}_{Brun\_in} &= \mathcal{L}_{\mathrm{sup}} + \mathcal{L}_{\mathrm{dcp}} \\
 &= \sum_{k=1}^{K} \mathbb{E}_{(x^s,y^s) \sim \mathcal{D}_{S_k}} \mathcal{L}_{\mathrm{det}} (\phi(x^s), y^s) + \mathcal{L}_{\mathrm{dcp}},
\end{aligned}
\end{equation}
where $L_{\text{det}}$ represents the overall loss for object detection.

\textbf{Teacher-Student mutual learning stage}: Both the teacher and student models are initialized from the result of Burn-in stage. The teacher model applies weak augmentations to the IR target domain data $\mathcal{D}_t$ to generate pseudo-labels $\hat{y}^t$ as supervision signals. The student model receives strongly augmented IR data and is imposed with semi-supervised constraints $\mathcal{L}_{\mathrm{unsup}}$ from the pseudo-labels $\hat{y}^t$, while also keep being supervised by $\mathcal{L}_{\mathrm{sup}}$ and $\mathcal{L}_{\mathrm{dcp}}$. The loss of student model is defined as Eq. \ref{Eq:mutual}.
\begin{equation}
\label{Eq:mutual}
\begin{aligned}
    \mathcal{L}_{\mathrm{TS\_std}} = \mathcal{L}_{\mathrm{MT}} (\phi(x^t), \hat{y}^t) + \mathcal{L}_{\mathrm{sup}} + \mathcal{L}_{\mathrm{dcp}},
\end{aligned}
\end{equation}
where $\mathcal{L}_{\mathrm{MT}}$ represents the overall loss for Mean-Teacher \cite{liu2021unbiased} object detection.
The parameters of the teacher model are dynamically integrated with the optimization trajectory of the student model through the EMA strategy. Due to the large domain gap in RGB-IR adaptation, the student model undergoes significant fluctuations. To mitigate this, we introduce the Student Stabilization Module (SSM) \cite{varailhon2024source} to enhance stability and adaptation.
\begin{equation}
\begin{aligned}
   \text{EMA}: \quad \theta_t \leftarrow \alpha_{ema} \theta_t + (1 - \alpha_{ema}) \theta_s , \\
   \text{SSM}: \quad \theta_s \leftarrow \alpha_{ssm} \theta_s + (1 - \alpha_{ssm}) \theta_t .
\end{aligned}
\end{equation}
Here, $\theta_t$ represents the parameters of the teacher network, $\theta_s$ represents the parameters of the student network, and $\alpha_{ema}$, $\alpha_{ssm}$controls the retention ratio of EMA and SSM historical parameters. 


\subsection{SAID}
\label{sec.SAID}

\begin{figure*}[t]
    \centering
     \includegraphics[width=1\linewidth]{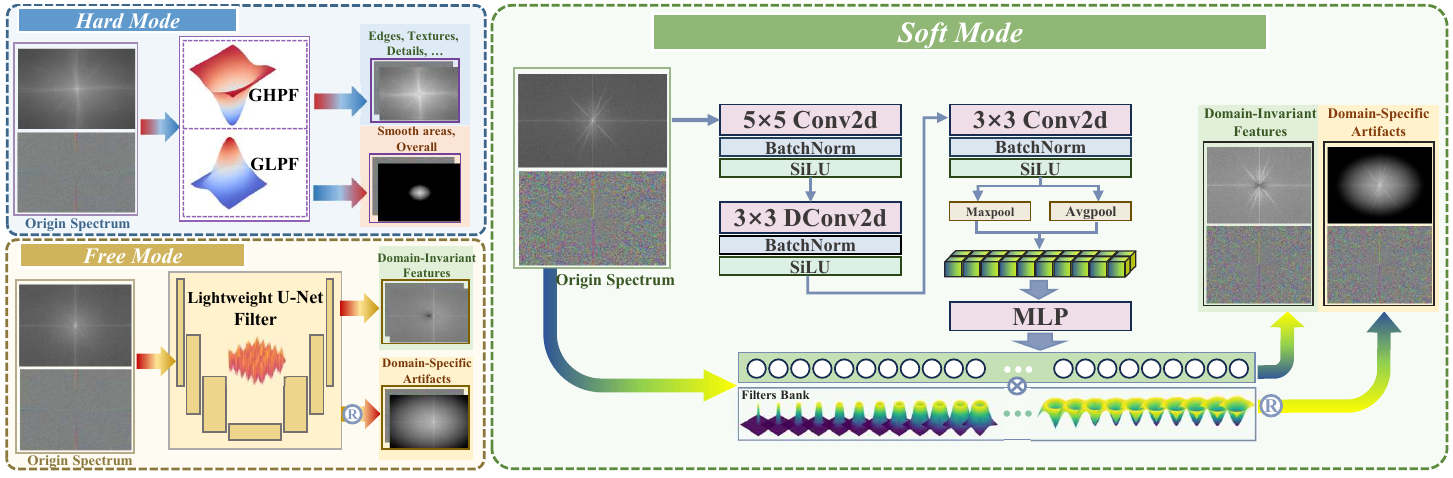}
  
     \caption{The SAID design includes: \textbf{Hard mode} uses a Gaussian low-pass filter (GLPF) and a Gaussian high-pass filter (GHPF). \textbf{Free mode} contains a lightweight U-Net. \textbf{Soft mode} is based on a Filter Bank design, combining the strengths of the first two modes.}
     \label{fig:SAID}
  \end{figure*}

To bridge the significant gap in adaptation from the RGB domain to the IR domain in real-world complex scenarios, we designed an Spectral Adaptive Idempotent filtering (SAID) module that leverages the frequency domain for information decoupling. The idempotent filtering module can decouple spectral information into DI information and DS interference. The fundamental design of the idempotent filtering layer is to satisfy $F^n(A)=F(A)$ \cite{shocher2023idempotent, xu2024idempotence}, which is constructed to meet idempotent invariance. SAID ensures decoupled representations across multiple data domains. To achieve this SAID filtering module design, we consider three different modes:

\textbf{SAID-Hard Mode}: In the hard filtering mode, we directly use a Gaussian high-pass filter as the filtering module to decompose the spectral information into low-frequency and high-frequency information. Its mathematical form is $\mathcal{F}_{\text{hard}}(A) = A \odot \left(1 - e^{-D^2 / 2\sigma_h^2}\right)$, where \( D \) is the radial spectral distance, $\sigma_h$ controls the cutoff frequency. The Semi-Group Property of Gaussian functions inherently ensures that the filter is idempotent.

\textbf{SAID-Soft Mode}: Mutually coupled filter bank with $N$ filters is designed to implement a dynamic filter. This design is inspired by the circular distribution of frequency domain and biological vision. By constructing a normalized frequency domain coordinate system $(X(u)=(u-c_{\mathrm{col}})/{\max{(h,w)}},Y(v)=(v-c_{\mathrm{row}})/{\max{(h,w)}})$, we define a radial frequency distance field $D(u,v)=\sqrt{X^{2}+Y^{2}}$. Here, \( u \) and \( v \) are the spectral domain coordinates, \( c_{\text{col}} \) and \( c_{\text{row}} \) are the coordinates of the center point, and \( h \) and \( w \) are the height and width of the image, respectively. A dynamic bandwidth allocation strategy $\Delta f=0.5/(N+1)$ is adopted to generate continuous pass-bands covering the Nyquist frequency range \cite{landau1967sampling} $[0,0.5]$. Each band-pass filter is constructed using a Gaussian difference operator as Eq. \ref{Eq:Gaosi}.
\begin{equation}
\label{Eq:Gaosi}
G_k(D)=\exp\left(-\frac{D^2}{2\widetilde{\sigma}_n^2}\right)-\exp\left(-\frac{D^2}{2\sigma_n^2}\right),\end{equation}
where $\sigma_{n}=n\Delta f$ and $\widetilde{\sigma}_{n}=(n+1)\Delta f$ control the pass-band boundaries, forming frequency-domain basis functions with smooth transition characteristics. This design ensures: \textbf{Full frequency coverage}:The continuous pass-bands cover the entire Nyquist frequency range $[0,0.5]$. \textbf{Heisenberg-optimal time-frequency locality}: The Gaussian window function achieves the optimal time-frequency balance, minimizing the Heisenberg uncertainty \cite{busch2007heisenberg} and avoiding spectral leakage. 

We feed the amplitude and phase spectrum into a convolutional flow to get a spectrum feature, while maintaining controllable parameters. The feature regresses a weight vector through the network in Figure \ref{fig:SAID}. The weight vector combines the filter bank into an adaptive spectral filter $H_{\text{inv}}(u,v)$ for extracting DI features. The complementary filter $H_{\text{spe}}(u,v)=1-H_{\text{inv}}(u,v)$ is used to extract DS features.

\textbf{SAID-Free Mode}: We use a lightweight U-Net \cite{ronneberger2015u} to regress a fully adaptive filter $H_{\text{inv}}$ from spectrum domain. This U-Net utilizes depthwise separable convolutions \cite{chollet2017xception} and channel attention mechanisms \cite{wang2020eca}, which develop spectrum modeling capabilities while maintaining a controllable number of parameters. The filter $H_{\text{spe}}$ is created using the same design as the soft mode.

\textit{To guarantee that the soft mode and free mode of SAID also exhibit idempotent filtering capabilities, we adopt a self-distillation-driven decoupling strategy.} Let $\text{DI}$ and $\text{DS}$ denote the spectrum decoupled by SAID. The spectrum of DI information filtered by the first SAID is reprocessed by a parameter-shared second SAID filter to get the $\text{DI}'$ and $\text{DS}'$. In the same manner, we obtain $\text{DI}''$ and $\text{DS}''$ from $\text{DI}'$. On this basis, we construct an idempotent decoupling design.

The domain-invariant information should remains unchanged or exhibits only proportional differences, which has strong positive correlation from each other. And the domain-specific interference, already removed in the first SAID filtering, should yield relatively independent results in three SAID stages. We use a Pearson Correlation Coefficient (PCC) \cite{HAHSVAUGHN2023734} -based spectral decoupling loss:
\begin{equation}\mathcal{L}_{\mathrm{dcp}}=\frac{\left(\mathrm{PCC}(DS,DS')+\mathrm{PCC}(DS,DS'')\right)^2}{\left(\mathrm{PCC}(DI,DI')\cdot\mathrm{PCC}(DI,DI'')\right)^k+\epsilon},\end{equation}
where $\mathrm{PCC}(X,Y)={\mathrm{cov}(X,Y)}/{\sigma_X\sigma_Y}$, $\mathrm{cov}(X,Y)=\frac{1}{N}\sum_i(X_i-\mu_X)(Y_i-\mu_Y)$, $\sigma_X=\sqrt{\frac{1}{N}\sum_i(X_i-\mu_X)^2}$. The numerator forces DS interference to become uncorrelated after multiple decoupling iterations. The denominator encourages DI information to remain highly correlated before and after decoupling. $k$ controls constraint strength, and $\epsilon$ prevents division by zero.

Based on the above design, during the Burn-in stage, SAID leverages the DI and DS among multiple source domains to form a strong decoupling ability in spectral domain, that makes the initialized model less sensitive to domain bias when generating pseudo-labels in the next stage. In the Teacher-Student mutual learning stage, the IR target domain receives higher-quality pseudo-labels to prevent model collapse, enhancing unsupervised domain adaptation on the IR target domain.

\subsection{SS-Coupling}
\label{sec.Couple}

SS-Coupling is proposed to couple the DI and DS decoupled by SAID with the features $\{F_B^{(l)} \}_{l=1}^5$ extracted by the spatial backbone. During the coupling process, $F_{inv}^{(l)}$ is adjusted through inverse Fourier transform and convolutional flows to fit the dimensions of the spatial features from DI, constructing spectral feature pyramid. Based spatial features contain spatial correlation information in the earlier layers, while the features in the later layers possess semantic information, we propose the SS-Coupling strategy:

For the features of earlier layers ($l=1, 2$), we use $F_{inv}^{(l)}$ to construct a attention map $A_{inv}^{(l)}$ for local feature selection. Spectral DI guidance enhances the spatial DI features.
\begin{equation}
\label{eqRes}
    F_{out}^{(l)} = \alpha_l \cdot A_{inv}^{(l)} + (1 - \alpha_l) \cdot F_B^{(l)},
\end{equation}
where $F_B^{(l)}$ denotes the $l-th$ feature of the backbone network and $\alpha_l$ is the adaptive fusion coefficient.

For the features of later layers ($l = 3, 4, 5$), cross-attention \cite{huang2019ccnet} is utilized to couple the $F_{inv}^{(l)}$ and $F_f^{(l)}$.
\begin{equation}
\resizebox{\columnwidth}{!}{$
\begin{aligned}
    Q^{(l)}, &K^{(l)}, V^{(l)} = Ln(W_Q F_f^{(l)}), Ln(W_K F_B^{(l)}), Ln(W_V F_B^{(l)}), \\
    F_{out}^{(l)} &= \alpha_l \cdot softmax\left(\frac{Q^{(l)} (K^{(l)})^T}{\sqrt{d}}\right) V^{(l)} + (1 - \alpha_l) \cdot F_B^{(l)}, 
\end{aligned}
$}
\end{equation}
where $Ln(:)$ denotes layer normalization and $d$ represents the feature dimension. This achieves the coupling of DI information with the spatial features.

To address the dynamic suppression requirements of DS interference, we use a guidance mechanism. To constructs DS tokens directly from the spectral DS features $F_{spe}$, We design a lightweight embedding module $G:\mathcal{R}^{2\times H\times W} \rightarrow \mathcal{R}^{N_t\times d}$, which compresses the dimensions through convolutional layers and pooling layers and maps the features to multiple guidance tokens for the DETR encoder through linear layers. Each token contains guidance information across different aspects. $T_{spe} = G(F_{spe})$ is used to guide the model in excluding DS interference during the process of transforming spatial tokens into bounding boxes, mitigating the impact of significant domain gaps.

\section{Experiments}

\subsection{Datasets and Evaluation Protocols}

We conduct our experiments on three datasets, including a new designed evaluation protocols from RGB to IR domains:

\textbf{FLIR-ADAS \cite{farooq2021object}:} In our research, to evaluate the adaptation ability of the real and complex distribution of the RGB domain to the IR domain, we chose the FLIR-ADAS dataset as the basis for construction. Compared to the FLIR dataset, FILR-ADAS contains wider data distribution, such as daytime, nighttime, and glare conditions. Unlike the FLIR dataset, which shares annotations between RGB and IR images, the FLIR-ADAS dataset includes 9,233 independently annotated RGB images and 9,711 annotated IR images without aligned, covering 15 different object categories. We focused only on objects such as “person”, “bike”, “car”, “sign”, and “light”, which have complete labels, to ensure the accuracy of our evaluation.

\textbf{FLIR \cite{FLIR}:} A dataset of real-world urban street scenes contains includes 5,142 precisely aligned pairs of color and IR images, comprising 4,129 training images and 1,013 validation images with pixel-level annotations.

\textbf{RGH$\rightarrow$IR FLIR-ADAS evaluation:}
Based on the FLIR-ADAS dataset, we conducted operations to ensure the authenticity of the evaluation from RGB to IR domain adaptation, such as eliminating weak alignment and preserving the diversity of the source domain. We used 5,663 labeled RGB images as the source domain training set, 4,856 unlabeled IR images as the target domain training set, and 1,144 IR images as the target domain evaluation set. This dataset ensures the scientific nature of the annotations, prevents information leakage caused by weak alignment, thereby enhancing the effectiveness of evaluating the RGB-to-IR domain adaptation task. See the Appendix for more details.

\textbf{RGH$\rightarrow$IR FLIR evaluation:}
We follow D3T \cite{do2024d3t} for the same FLIR dataset settings. The RGB training dataset contained 2,064 labeled images and the IR target domain training dataset contains 2,064 IR unlabeled images. And 1,013 IR labeled images for evaluation purpose.

\subsection{Implementation Details}

In accordance with the baseline settings, we equipped the
Transformer detector with a ResNet-50 \cite{he2016deep} backbone pre-trained on ImageNet \cite{deng2009imagenet} in our experiments. Our experiments were conducted using 2 NVIDIA L40 GPUs with a batch size of 8. For data augmentation and test size, we adopted the same augmentation strategy as previous works \cite{deng2023harmonious,farooq2021object,chen2025datr}, and resized the shortest edge of the images to a maximum of 800 pixels. For the FLIR-ADAS dataset, we reproduced the previous UDAOD methods. We conducted experiments with different SSM \cite{varailhon2024source} parameters, EMA \cite{hunter1986exponentially} strategy parameters, SAID mode choices, decoupled loss constraint strength, decoupled loss coefficient, and ablation study. Each experiment consisted of 20k iterations and was repeated three times to obtain the average result. For the FLIR datasets, we followed the D3T \cite{do2024d3t} settings for comparison experiments. More details are provided in the Appendix.

\subsection{Main Results}

\begin{table*}[ht]
  \caption{Results of adaptation from RGB$\rightarrow$IR FLIR-ADAS. In the model design: A - Semi-supervised (Mean-Teacher), B - Unsupervised Domain Adaptation, C - RGB$\rightarrow$IR Unsupervised Domain Adaptation. “Source only” and “Oracle” refer to the DINO trained by only using labeled source domain data and labeled target domain data, respectively. The mAP in the table are calculated according to the COCO style. The best results are marked with bold, the sub-best results are marked with an underline.}
  \label{tab:FLIR-ADAS}
  \centering
    \begin{tabular}{l|c|c|ccccc|c}
    \toprule
    Method & Reference & \makecell{Model Design \\ 
  \begin{tabular}{c|c|c}  
    A & B & C
  \end{tabular}
}  & Person & Bicycle  & Car   & Sign  & Light & mAP \\
    \midrule
    \midrule
    \multicolumn{9}{l}{CNN type frameworks:} \\
    \midrule
    Efficient Teacher \cite{xu2023efficient} & arXiv'23 & \makecell{
    \begin{tabular}{ccc} $\checkmark$ & $\quad$ & $\quad$
    \end{tabular}} & 48.52 & 38.57 & 63.33 & 24.25 & 5.67  & 36.07 \\
    HT \cite{deng2023harmonious} & CVPR'23   &  \makecell{
    \begin{tabular}{ccc} $\checkmark$ & $\checkmark$ & $\quad$
    \end{tabular}} & \underline{54.19} & \textbf{41.97} & 70.89 & 29.05 & 16.48 & 42.52 \\
    D3T \cite{do2024d3t} & CVPR'24 &  \makecell{
    \begin{tabular}{ccc} $\checkmark$ & $\checkmark$ & $\checkmark$
    \end{tabular}}  & 52.75 & \underline{41.87} & \underline{71.78} & 30.99 & \underline{17.80}  & \underline{43.04} \\
    \midrule
    \multicolumn{9}{l}{Transformer type frameworks:} \\
    \midrule
    \textbf{Source only} (DINO \cite{zhang2022dino}) & ICLR'23 &  & 39.32 & 32.53 & 57.97 & 20.05 & 9.20 & 31.81 \\
    \textbf{Oracle} (DINO \cite{zhang2022dino}) & ICLR'23 &  & 77.05 & 45.65 & 82.51 & 56.17 & 59.57 & 64.19 \\
    \midrule
    MixPL-DINO \cite{chen2023mixed} & arXiv'23 & \makecell{
    \begin{tabular}{ccc} $\checkmark$ & $\quad$ & $\quad$
    \end{tabular}}  & 46.26 & 34.85 & 62.14 & 27.73 & 11.27 & 36.45 \\
    Semi-DETR \cite{zhang2023semi} & CVPR'23 & \makecell{
    \begin{tabular}{ccc} $\checkmark$ & $\quad$ & $\quad$
    \end{tabular}} & 51.40  & 36.83 & 70.68 & 32.12 & 15.24 & 41.25 \\
    Remote Sensing Teacher \cite{han2024remote} &TGRS'24 & \makecell{
    \begin{tabular}{ccc} $\checkmark$ & $\checkmark$ & $\quad$
    \end{tabular}} & 43.91 & 33.64 & 64.36 & 32.72 & 16.17 & 38.16 \\
    DATR \cite{chen2025datr}& TIP'24 & \makecell{
    \begin{tabular}{ccc} $\checkmark$ & $\checkmark$ & $\quad$
    \end{tabular}} & 52.38 & 38.13 & 69.59 & \underline{32.84} & 16.94 & 41.98 \\
    \rowcolor[gray]{0.92}SS-DC (ours) & / & \makecell{
    \begin{tabular}{ccc} $\checkmark$ & $\checkmark$ & $\checkmark$
    \end{tabular}}  & \textbf{56.36} & 41.58 & \textbf{75.82} & \textbf{39.83} & \textbf{26.51} & \textbf{48.02} \\
    
    \bottomrule
    \end{tabular}%
\end{table*}%

\begin{table}[ht]
  \caption{Results of UDAOD methods adaptation from RGB$\rightarrow$IR FLIR. “Source only” and “Oracle” refer to the DINO trained by only using labeled source domain data and labeled target domain data, respectively. The mAP in the table are calculated according to the VOC style. The best results are marked with bold, the sub-best results are marked with an underline.}
  \label{tab:FLIR}%
\setlength{\tabcolsep}{5pt} 
  \centering
  \small
  \begin{tabular}{c|c|>{\centering}m{0.7cm}>{\centering}m{0.7cm}>{\centering}m{0.7cm}|c} 
    \toprule
    \makecell{RGB$\rightarrow$IR} & Method & Person & Bicycle & Car & mAP \\
    \midrule
    \multirow{6}[0]{*}{\ding{55}} & \textbf{Source only} & 28.51 & 34.72 & 49.96 & 37.73 \\
          & SWDA \cite{li2023improving}  & 30.91 & 36.03 & 47.94 & 38.29 \\
          & EPM \cite{hsu2020epm}   & 40.97 & 38.95 & 53.83 & 44.60 \\
          & DATR \cite{chen2025datr}  & 68.02 & 46.22 & 74.88 & 63.04 \\
          & HT \cite{deng2023harmonious}    & \underline{70.87} & 48.11 & 78.45 & 65.81 \\
          & \textbf{Oracle} & 74.80 & 54.72 & 85.55 & 71.69 \\
    \midrule
    \multirow{1}[0]{*}{\ding{51}} & D3T \cite{do2024d3t}   & 70.77 & \underline{57.44} & \underline{79.68} & \underline{69.30} \\
    \rowcolor[gray]{0.92}\multirow{1}[0]{*}{\ding{51}} & SS-DC (Ours) & \textbf{74.25} & \textbf{61.21} & \textbf{81.84} & \textbf{72.43} \\
  \end{tabular}%
\end{table}%

\textbf{RGB$\rightarrow$IR FLIR-ADAS evaluation:}
The data in Table \ref{tab:FLIR-ADAS} shows that our SS-DC framework achieves significant performance in RGB-IR domain adaptation. In the FLIR-ADAS evaluation for adapting from the complex distribution of the RGB source domain to the IR target domain, our method's average precision is 6.35\% higher than the baseline Semi-DETR \cite{zhang2023semi}, 6.04\% higher than the novel DATR \cite{chen2025datr} algorithm, and 4.98\% higher than the D3T \cite{do2024d3t} algorithm which is the state-of-the-art in FLIR. 
SS-DC framework which utilizes the complex distribution within the RGB domain to enhance domain adaptation. The results fully demonstrate the effectiveness of our method in solving the domain gap from RGB to IR domains

\textbf{RGB$\rightarrow$IR FLIR evaluation:}
The data in Table \ref{tab:FLIR} proves that we still maintain our advanced position in the FLIR dataset test. We outperform D3T \cite{do2024d3t} algorithm by 3.13\%. So we argue that there is no completely uniform data domain distribution \cite{liu2024replicable}, and one macroscopic source domain does not mean there are no DI and DS to extract. The experiment results prove that our method remains effective even when the source domain is relatively uniform.

\subsection{Ablation Experiments}
We make ablations and detail discussions based on FLIR-ADAS dataset in this section.

\begin{figure}[t]
  \centering
   \includegraphics[width=1.0\linewidth]{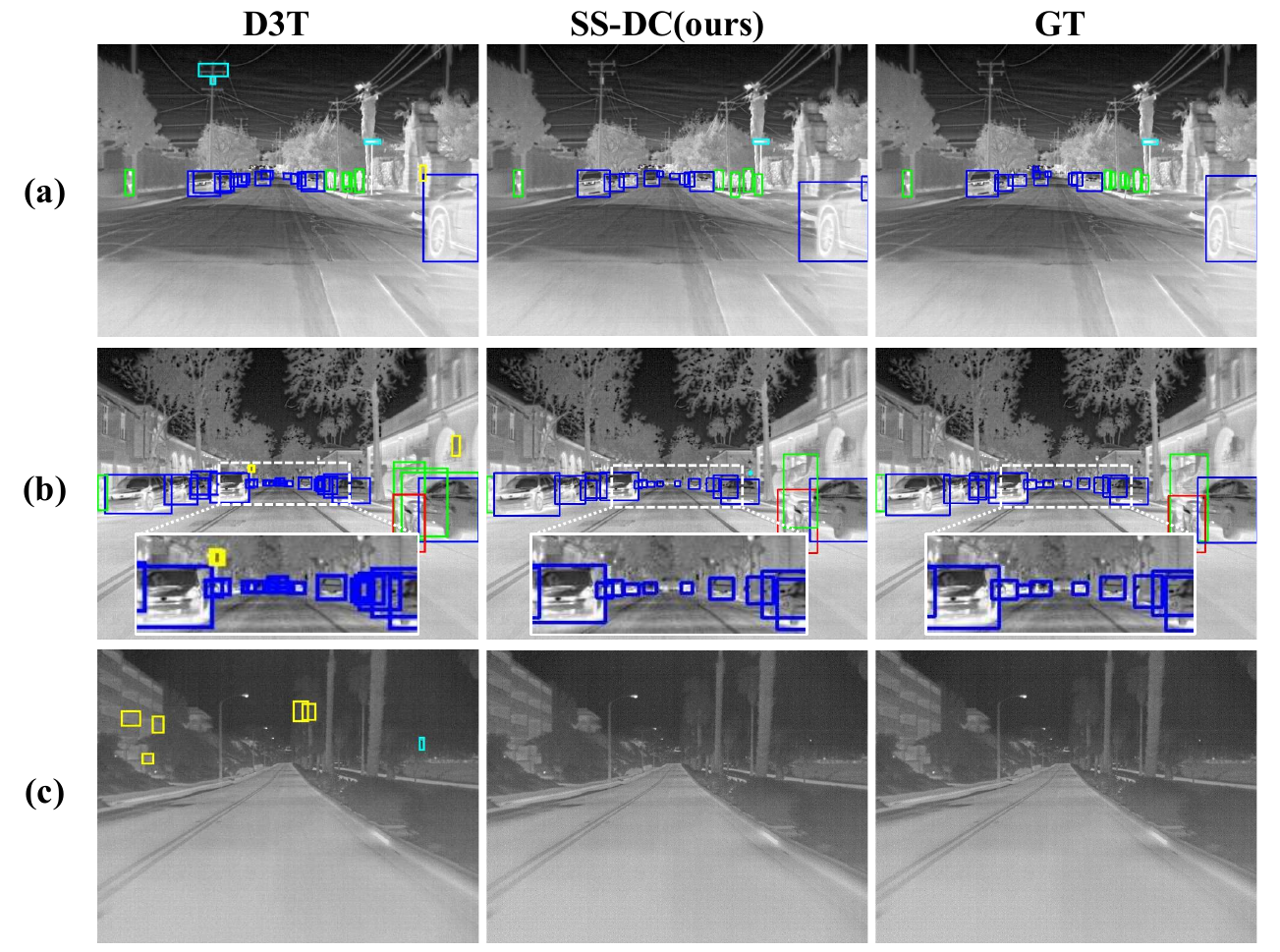}
   \caption{Visualization comparison of model inference results for UDAOD from RGB to IR domains includes D3T and SS-DC. (a), (b), and (c) represent the inference results of different images.}
   \label{fig:Visual}
\end{figure}

\textbf{Visualization:}
The visualization results show that SS-DC achieves superior performance in handling objects with blurred contours and objects with faint feature in IR images. And the error detection ratio is significantly lower than in previous method. SS-DC can better address the unique challenges of IR domain, including blur and significant domain shifts, as shown in Figure \ref{fig:Visual}.

\textbf{Ablation Study:}
As shown in Table \ref{tab:Ablat}, the ablation study of SS-DC evaluates the contributions of each component of SS-DC framework on the FLIR-ADAS dataset. Starting from a Semi-DETR \cite{zhang2023semi} baseline with an average precision of 41.25\%. The mAP is enhanced to 45.16\% after incorporating the SAID module to decouple DI and DS information and combining DI with spatial shallow features. Based on the strong decoupling ability of SAID and the coupling capability of SS-Coupling boost the mAP to 46.98\%. The application of more parameter adjustments and the student stabilizer module elevate the model's mAP to 48.02\%.

\begin{table}[htbp]
 \caption{Ablation studies on RGB$\rightarrow$IR FLIR-ADAS for UDAOD SS-DC framework. The best result is marked with bold.}
  \label{tab:Ablat}%
\small
\setlength{\tabcolsep}{12pt} 
  \centering
  \small
    \begin{tabular}{ccc|c}
    \toprule
    SAID  & SS-Coupling & SSM   & mAP \\
    \midrule
    \midrule
          &       &       & 41.25 \\
    $\checkmark$     &       &       & 45.16 \\
    $\checkmark$     & $\checkmark$     &       & 46.98 \\
    $\checkmark$     & $\checkmark$     & $\checkmark$     & \textbf{48.02} \\
    \bottomrule
    \end{tabular}%

\end{table}%

\textbf{SAID Mode Comparison:}
Experiments comparing the hard, soft, and free modes of SAID, as designed in Sec. \ref{sec.SAID}, show that the soft mode achieves the best training result, with a mAP 48.02\%. As shown in Table \ref{tab:SAID}, the hard mode, with its inherent decoupling advantages, achieve an mAP of 46.31\%, while the free mode which is commonly used in precious works reaches 47.15\%. Due to the circular distribution of the spectral domain, the free mode, with its limited parameters, cannot fully fit the characteristics of the spectral distribution. The hard mode cannot perform adaptive spectral processing for different domains. Therefore, the soft mode processing paradigm effectively integrates the advantages of both hard and free modes, emerging as an innovative and effective new paradigm for frequency domain processing. 

\begin{table}[htbp]
  \caption{Ablation studies on RGB$\rightarrow$IR FLIR-ADAS for SAID Mode Comparison. The best result is marked with bold.}
  \label{tab:SAID}%
\setlength{\tabcolsep}{8pt} 
  \centering
    \begin{tabular}{c|ccc}
    \toprule
          & Hard mode & Soft mode & Free mode \\
    \midrule
    \midrule
      mAP   & 46.31 & \textbf{48.02} & 47.15 \\
    \bottomrule
    \end{tabular}%
\end{table}%

\begin{figure}[t]
  \centering
   \includegraphics[width=1.0\linewidth]{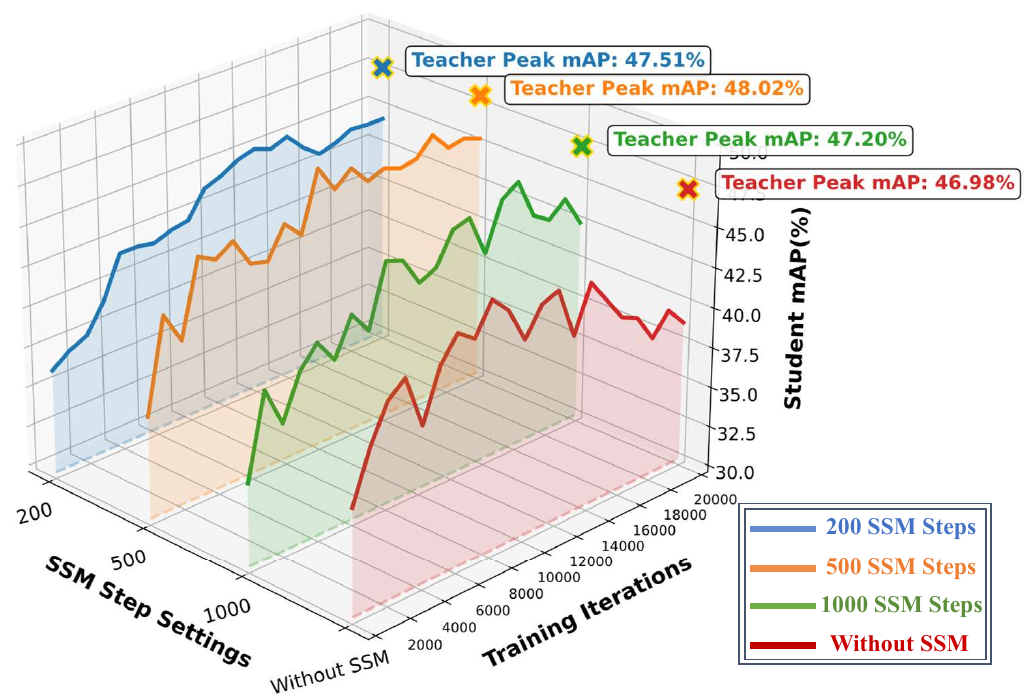}

   \caption{Ablation studies on RGB$\rightarrow$IR FLIR-ADAS for  different Student Stabilisation Module Step.}
   \label{fig:SMM}
\end{figure}

\textbf{Student Stabilizer Effectiveness:}
Inspired by prior work \cite{varailhon2024source}, we recognize that instability of the student model often leads to failure in the Mean Teacher paradigm, especially in the challenging task of adapting from RGB to IR domains. Therefore, we conducted effectiveness tests on the introduced SSM strategy with $\alpha_{ssm}=0.5$. As shown in Figure \ref{fig:SMM}, SSM represents a trade-off between training stability and the exploratory nature of the student model. When the SSM step size is set to 500 iterations, it significantly enhances the domain adaptation process.

\textbf{Filter Bank Fine-Grained:}
The results of experiments on different preset numbers of filter banks, as shown in Table \ref{tab:Filterbank}, indicate that a low number of filters (num=10) limits the model's adaptability due to insufficient degrees of freedom. Conversely, a high number of filters (num=500) makes it difficult for the model to achieve good parameter weight regression. The optimal performance reaches 48.02\% in the tests.

\begin{table}[htbp]
  \caption{Ablation studies on RGB$\rightarrow$IR FLIR-ADAS for Filter Bank Fine-Grained. The best result is marked with bold.}
  \label{tab:Filterbank}%
\small
\setlength{\tabcolsep}{3pt}
  \centering
    \begin{tabular}{c|cccccc}
    \toprule
        \makecell{Number of Filters \\in Filter Bank} & 10 & 20 &50 &100 & 200 &500\\
    \midrule
    \midrule
    mAP   & 45.10 & 46.82 & 47.40 & \textbf{48.02} & 46.28 & 46.43\\
    \bottomrule
    \end{tabular}%
\end{table}%

\textbf{Self-Distillation Decoupling Loss Parameter:}
We tested different constraint strengths \( k \) and loss coefficients, as shown in Table \ref{tab:losspara}. Through various parameter combinations, we identified the optimal performance point for this loss function.

\begin{table}[htbp]
  \caption{Ablation studies on RGB$\rightarrow$IR FLIR-ADAS for Self-Distillation Decoupling Loss Parameter. The best result is marked with bold.}
  \label{tab:losspara}%
\small
  \centering
    \begin{tabular}{c||ccc}
    \toprule
    mAP of different  \( k \) \& $\lambda_{dcp}$ &  \( k \)=1   &  \( k \)=2   &  \( k \)=3 \\
    \midrule
    \midrule
    $\lambda_{dcp}$=5 & 45.63  & 45.21  & 44.27  \\
    $\lambda_{dcp}$=10 & 46.30  & 47.14  & 45.89  \\
    $\lambda_{dcp}$=50 & 47.39 & \textbf{48.02 } & 44.93  \\
    $\lambda_{dcp}$=100 & 47.42  & 46.11  & 42.20  \\
    \bottomrule
    \end{tabular}%
\end{table}%

More ablation experiment details are shown in the Appendix.

\section{Conclusion}

This paper presents the SS-DC framework to bridge the significant domain gap between RGB and IR domains for UDAOD. The SAID module decouples DI and DS information via a self-distillation-driven loss. The SS-Coupling strategy couples the spectral DI and DS information with spatial features in a multi-level, multi-structured way. We also carefully tune various methods and parameters to fit RGB-IR adaptation under the Mean Teacher paradigm. Additionally, we propose a new experimental protocol based on FLIR-ADAS dataset to more scientifically and realistically evaluate the RGB-IR UDAOD task. Experiment results demonstrate the effectiveness of our approach.

\end{document}